\RequirePackage{amsmath}
\documentclass[runningheads]{llncs}
\usepackage[T1]{fontenc}
\usepackage{graphicx}
\usepackage{color}

\usepackage{multirow}
\newcommand{\bftab}{\fontseries{b}\selectfont}

\DeclareMathOperator{\SSIMAE}{SSIMAE}

\usepackage{tikz}

\usepackage{amssymb}
\usepackage{xcolor}

\usepackage{placeins}

\newcommand{\figref}[1]{\figurename~\ref{#1}}
\newcommand{\tabref}[1]{\tablename~\ref{#1}}

\newcommand{\anoncmd}[2]{%
\ifdefined\anonymise%
#2%
\else%
#1%
\fi%
}

\newcommand{\reffcmd}[2]{%
\ifdefined\shortref%
#1%
\else%
#2%
\fi%
}

\reffcmd
{
\usepackage[backend=biber,
sorting=nyt,
url=false,
doi=false,
maxnames=6,
minnames=6,
style=lncs]{biblatex}
\DeclareFieldFormat{titlecase}{\MakeSentenceCase*{#1}}
\addbibresource{library.bib}
}
{
}

\usepackage[hidelinks]{hyperref}

\begin{document}

\title{Transferring Relative Monocular Depth to Surgical Vision with Temporal Consistency}
\titlerunning{Transferring Relative Monocular Depth to Surgical Vision}

\anoncmd
{
\author{Charlie Budd\inst{1}
\and Tom Vercauteren\inst{1,2}}
\authorrunning{C. Budd et al.}
\institute{
King’s College London, Biomedical Engineering \& Imaging Science, London
\and Hypervision Surgical Limited, 1st Floor 85 Great Portland Street, London
}
}
{
\author{anonymised}
\institute{anonymised}
}
\maketitle

\begin{abstract}
Relative monocular depth, inferring depth 
up to 
shift and scale from a single image, is an active research topic.
Recent deep learning models, trained on large and varied meta-datasets, now provide excellent performance in the domain of natural images.
However, few datasets exist which provide ground truth depth for endoscopic images, making training such models from scratch unfeasible.
This work investigates the transfer of these models into the surgical domain, and presents an effective and simple way to improve on standard supervision through the use of temporal consistency self-supervision.
We show temporal consistency significantly improves supervised training alone when transferring to the low-data regime of endoscopy, and outperforms the prevalent self-supervision technique for this task.
In addition we show our method drastically outperforms the state-of-the-art method from within the domain of endoscopy.
We also release our code, models, and ensembled meta-dataset, \texttt{Meta-MED}, establishing a strong benchmark for future work.
\end{abstract}

\section{Introduction}

\begin{figure}[t]
\centering
\includegraphics[width=0.7\textwidth]{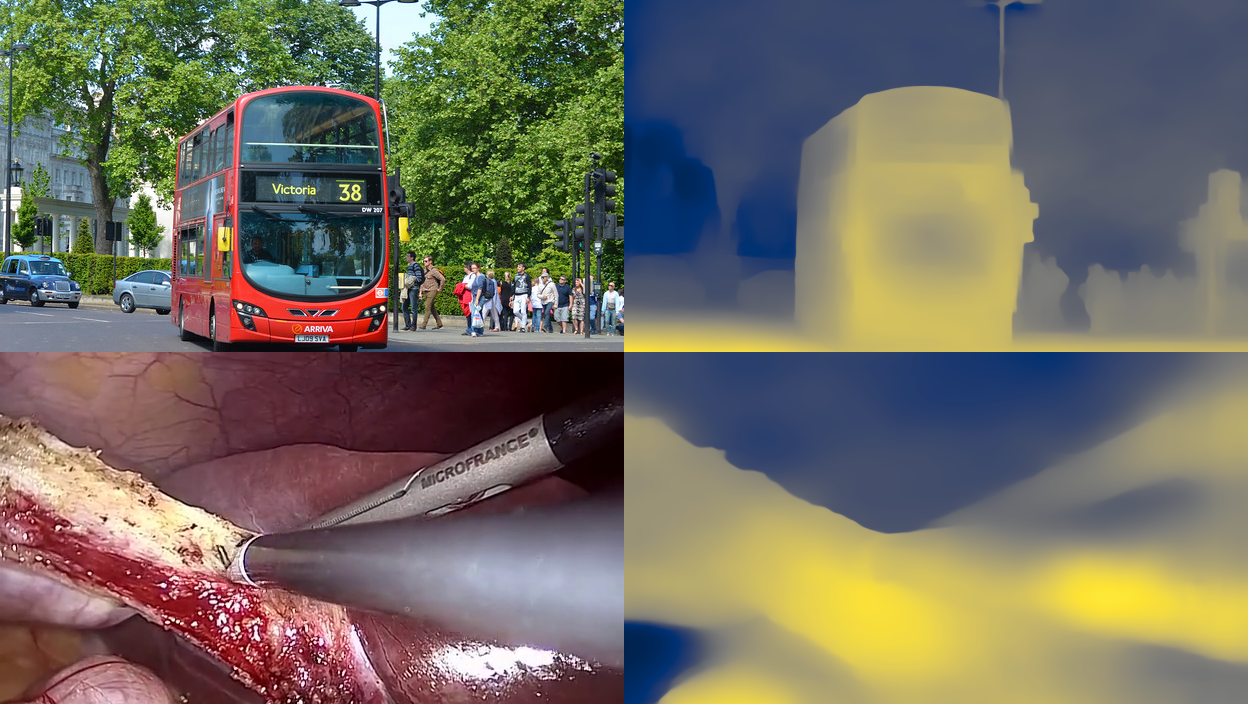}
\caption{Example images from the natural image and endoscopic domains with corresponding inverse relative depth-maps. The depth-maps are generated using a MiDaS model pre-trained on natural images. The ability of the model to transfer to the endoscopic domain is notable and speaks to the fundamental nature of the depth estimation task. However, closer inspection reveals significant flaws in the estimated depth-map.}
\label{depthmapexamples}
\end{figure}

%
The task of monocular depth in computer vision
focuses on estimation
from a single viewpoint, as opposed to using
multi-view geometry.
Example depth maps are shown in~\figref{depthmapexamples}.
Throughout this work, we focus on dense depth, wherein the depth is estimated for every pixel in the image, and relative depth, meaning the estimated depth values are only 
sought up to 
scale and shift.
Targeting relative depth,
as proposed in~\cite{Ranftl2022}, allows training on data from multiple sources without concern for  calibration or large expected range differences between applications.
State-of-the-art methods 
for natural images have exploited this property
to leverage huge \emph{meta-datasets}, composed of multiple existing datasets, to train large transformer based models.
The most recent MiDaS models~\cite{birkl2023midas}, the culmination of a series of recent works~\cite{Ranftl2021,Ranftl2022}, are trained on 1.4 million labelled images.
Depth Anything~\cite{yang2024depth}, released during the later stages of our work, iterates on MiDaS and trains
on 1.5 million labelled images and 62 million unlabelled images.

An adequate solution to monocular depth in surgery would act as a key enabler for many surgical vision tasks.
It would be directly useful for tasks such as intraoperative autofocus~\cite{Budd_2023}, AR overlays~\cite{RAMALHINHO2023102943}, and surgical site mapping~\cite{9478277}.
More generally, though, the ability of a model to correctly discern depth from an endoscopic image demonstrates a nuanced understanding of the geometry of a surgical scene, which may be key for better solving more complex vision tasks.
However, as very few surgical datasets with ground truth depth exist,
following the same approach as state-of-the-art natural image models is unfeasible.
Experimenting with these models, we observed that one of the main failure modes when inferring outside their training domain, and specifically in surgery, seems to be an uncertainty in the depth ordering of semantic elements.
This leads to temporal inconsistent predictions, an aspect noticed by other works~\cite{wang2023neural}.
We therefore choose to investigate the transfer of these models to the surgical domain while following the paradigm of ensembled meta-datasets. We focus our attention on temporal consistency which provides a path for self-supervised training.

%
In the natural image domain,
prior to the advent of current large datasets and models, structure from motion (SFM)
approaches had been used to generate pseudo ground truth depth~\cite{zhou2017unsupervised,luo2020consistent}.
This line of self-supervision approaches still dominates monocular depth in endoscopy~\cite{lou2024wssfmlearner,shao2022self,8889760,9145848}.
Due to the dynamic nature of surgery and the highly deformable nature of tissue and articulated surgical tools,
these methods are restricted to training on videos specifically collected for such tasks, featuring static tissues with little to no tool interaction.
To improve results in gastro-endoscopy, geometric prior knowledge have been exploited~\cite{YANG2023105989}.
While elegant, this does not provide a scalable solution for general surgical vision.
%
To the best of our knowledge, the state-of-the-art in general surgical vision with released models is AF-SFMLearner~\cite{shao2022self}.
An incremental improvement over this method
has been reported with WS-SFMLearner~\cite{lou2024wssfmlearner} 
but this model has not been publicly released.
Beyond SFM-based methods, in \cite{cui2024surgicaldino} the authors 
proposed to use large pretrained natural image models
as a starting point for endoscopy. They fine-tune DinoV2's depth estimation model~\cite{oquab2024dinov2} to various individual endoscopy depth datasets.
Due to the small sizes and low variability of the datasets used, generalisability is a natural concern. As they don't release the models, and present results for 
depth up to scale only rather than scale and shift,
benchmarking against their reported results is not possible.

In this work, we start by compiling \texttt{Meta-MED}, the first meta-dataset intended to train and evaluate relative monocular depth models in the endoscopic domain.
We then demonstrate the ability of large state-of-the-art relative depth models for natural images to generalise reasonably to our domain and that standard supervised fine tuning on our meta-dataset
provides a strong boost to performance.
We also experiment with adding self-supervision methods to the training pipeline to great effect, and demonstrate that our novel temporal consistency self-supervision outperforms a typical self-supervision approach.
We show that many of our models outperform our domain-specific state-of-the-art baseline AF-SFMLearner.
Finally, we release our code\footnote{https://github.com/charliebudd/transferring-relative-monocular-depth-to-surgical-vision}, models, and dataset, establishing a strong benchmark for future research.

\section{Materials and Methods}

\subsubsection{Datasets}
%
Two high-quality datasets for dense depth estimation in minimally invasive surgery are the SCARED~\cite{allan2021stereo} and SERV-CT~\cite{Edwards_2022} datasets.
The ground truth for these two datasets are collected using two different measuring techniques. SCARED uses structured light projection with a stereo camera to create sparse depth maps, whereas SERV-CT creates dense depth maps using a pre-operative CT which is registered to image space. Both of these datasets are small, consisting of just 45 and 16 images captured in 9 and 2 ex-vivo porcine samples, respectively.
While SCARED also has approximate depth maps for more frames, these are made by reconstruction between the ground truth frames.
We ignore these to reduce errors and redundancy in the data.
Due to the small size of these datasets, we choose to combine these to form a holdout testing dataset.

Our first fine-tuning approach relies on supervision with pseudo ground truth.
We make use of an ensemble of stereo endoscopic datasets with depth maps calculated using stereo disparity, a full list of which can be seen in \tabref{datatable}.
The stereo data totals a little over 3 hours of footage and is drawn from 57 surgeries, most of which were in-vivo porcine.
The videos are first sampled to one frame per second to reduce redundancy in the data.
The left and right images are then stereo rectified and cropped to remove borders where needed.
RAFT~\cite{teed2020raft}, a capable and easily available optical flow model which has seen previous usage in surgical videos
~\cite{Kiyasseh2023-ge,SESTINI2023102751}
, is then used to calculate the bidirectional optical flow maps $F_{a\rightarrow b}$ and $F_{b\rightarrow a}$.
In order to only provide supervision for regions of the image with a confident ground truth, we 
mask out correspondences where the bidirectional flow does not form a closed loop.
We define a warping operator $F \boldsymbol{\cdot} I$ as the image $I$ warped by the flow $F$.
We then define a correspondence mask as
\begin{equation}
    C_{a\rightarrow b} = \mathbf{1}(|F_{a\rightarrow b} + F_{a\rightarrow b} \boldsymbol{\cdot} F_{b\rightarrow a}| < \varepsilon)
    \label{eq-mask}
\end{equation}
where $\mathbf{1}$ is the indicator function
and $\varepsilon$ is 
set to 2 pixels for our experiments.
As the images are rectified,
the horizontal component of $F_{a\rightarrow b}$ gives us the stereo disparity and
the vertical component of the flow should be zero.
We use a threshold of 2 pixel displacement to allow some leniency but mask out the deviations.

The data described so far is combined to form, \texttt{Meta-MED}, the meta monocular endoscopic depth dataset. Due to combining disparate data sources, and maintaining strict splits, we believe this to be much more appropriate for training generalisable depth models than what is currently available.
However, the amount of training data is still relatively tiny, totalling just 14,310 images with pseudo ground truth depth, compared to 1.4 million used to train the MiDaS models.
Self supervision
allows us to leverage data that has no ground truth,
thereby learning
from a much broader domain during training.
In addition to \texttt{Meta-MED}, we choose to combine RobutsMIS~\cite{ross2020robust} and Cholec80~\cite{twinanda2016endonet} giving us a total of 78 hours of 25 fps footage (117k images) from 110 in-vivo human surgeries. The videos are highly dynamic and feature extensive tissue-tool interaction. Most of the videos in these datasets feature prominent endoscopic content areas, which are detected
and cropped out using the method in~\cite{budd2023rapid}.

\begin{table}[tb]
\centering
\caption{A tabulation of all the datasets used in this work. The domain column summarises the type of data, specifically, how many samples/surgeries, whether they were in-vivo or ex-vivo, and whether they were porcine, human, or mixed. The usage column indicates in what way the data was used.}
\label{datatable}
\begin{tabular}{|c|c|c|c|c|}
\hline
Source dataset & Ground truth & Format & Domain & Usage \\
\hline
RobutsMIS~\cite{ross2020robust} & \multirow{2}{*}{None} & 27h, 25 fps & 30 in-vivo human & \multirow{2}{*}{Self-supervision}\\
Cholec80~\cite{twinanda2016endonet} & & 51h, 25 fps & 80 in-vivo human & \\
\hline
EndoVis2017~\cite{allan20192017} & \multirow{4}{*}{Stereo disparity} & 50m, 1 fps & 10 in-vivo porcine & \multirow{4}{*}{Supervision}\\
EndoVis2018~\cite{allan20202018} & & 95m, 1 fps & 19 in-vivo porcine & \\
KidneyBoundary~\cite{Hattab2020} & & 12m, 1 fps & 15 in-vivo porcine & \\ 
StereoMIS~\cite{hayoz2023learning} & & 45m, 1 fps & 6 in-vivo mixed & \\ 
\hline
Hamlyn~\cite{ye2017selfsupervised} & Stereo disparity & 36m, 1 fps & 7 in-vivo mixed & Validation \\
\hline
SERV-CT~\cite{Edwards_2022} & Registered CT & 16 images & 2 ex-vivo porcine & \multirow{3}{*}{Testing} \\
SCARED~\cite{allan2021stereo} & Structured light & 45 images & 9 ex-vivo porcine & \\
SCARED Clips~\cite{allan2021stereo} & None & 23m, 25 fps & 9 ex-vivo porcine & \\
\hline
\end{tabular}
\end{table}

\subsubsection{Fine-tuning Losses}
Here, we introduce the three losses we work with in our experiments: standard supervision ($\mathcal{L}_{sup}$), and our self-supervision losses, including augmentation consistency ($\mathcal{L}_{aug}$) and temporal consistency ($\mathcal{L}_{temp}$).
During self-supervision, by comparing the output of the model with an output of the same model for a different input, it is possible for the model to \emph{collapse} to a trivial solution that maximises the objective function, for example, outputting all zeros no matter the input.
A common way to help prevent collapse is to have the two branches use slightly different models.
This can be achieved by having one branch use a \emph{slow/teacher} model $M_{slow}$, which does not receive gradient updates and instead use weights calculated via an exponential moving average of the \emph{fast/student} model $M_{fast}$ which is used for the other branch.
We adopt this approach for both our self-supervision losses.

\subsubsection{Standard Supervision}
In order to construct loss functions and evaluation metrics for relative depth, we must be able to compare two depth maps in a way that is invariant to any scale and shift between the two.
MiDaS and Depth Anything take the absolute difference between the two depth maps, each normalised by subtracting the median and dividing by the standard deviation, with some minor modifications to account for outliers.
We found it produced better results to perform a linear least squares fitting of the predicted depth $\vec{d}$ to the ground truth depth $\vec{d^*}$, the latter being normalised as above to provide a consistent loss magnitude between different depth sources. We refer to this as Scale and Shift Invariant Mean Absolute Error (SSIMAE):
\begin{equation}
    \SSIMAE(\vec{d}, \vec{d^*}) = \frac{1}{N}\sum_{i=0}^{N}|(\alpha\vec{d} + \beta) - \vec{\hat{d^*}}|_i
    \label{eq-fit}
\end{equation}
where $N$ is the number of unmasked pixels in $\vec{d^*}$;
$\vec{\hat{d^*}}$ is the normalised version of $\vec{d^*}$;
and $\alpha$ and $\beta$ are found by a linear least squares fit of $\vec{d}$ to $\vec{\hat{d^*}}$ ignoring any masked pixels.
We then define our standard supervision loss $\mathcal{L}_{sup}$ as the SSIMAE between the depth map inferred by our fast model and the ground truth:
\begin{equation}
    \mathcal{L}_{sup} = \SSIMAE\Big(M_{fast}(\vec{I}), \vec{d^*}\Big)
\end{equation}

\subsubsection{Augmentation Consistency Self-supervision}
A common method of self supervision is to provide the model with a weakly and strongly augmented image. The output from the weakly augmented image may then be used as pseudo ground truth for the output from the strongly augmented image. This was originally proposed in FixMatch~\cite{sohn2020fixmatch} for classification tasks and has been adapted to segmentation in endoscopy~\cite{wei2023segmatch}.
We use our slow model to infer a depth map for an image from our unlabeled datasets.
We apply an augmentation transform
which uses colour jittering $A^{c}$ for the image and affine spatial transformations $A^{s}$ applied jointly to the image and inferred depth map to keep them aligned.
The fast model is then used to infer the depth map from the strongly augmented image.
Our augmentation consistency loss $\mathcal{L}_{aug}$ is defined as 
\begin{equation}
    \mathcal{L}_{aug} = \SSIMAE\Big(M_{fast}\big(A^c(A^s(\vec{I}))\big), A^s\big(M_{slow}(\vec{I})\big)\Big)
\end{equation}

\subsubsection{Temporal Consistency Self-supervision}
\begin{figure}[t]
\centering
\includegraphics[width=0.9\textwidth]{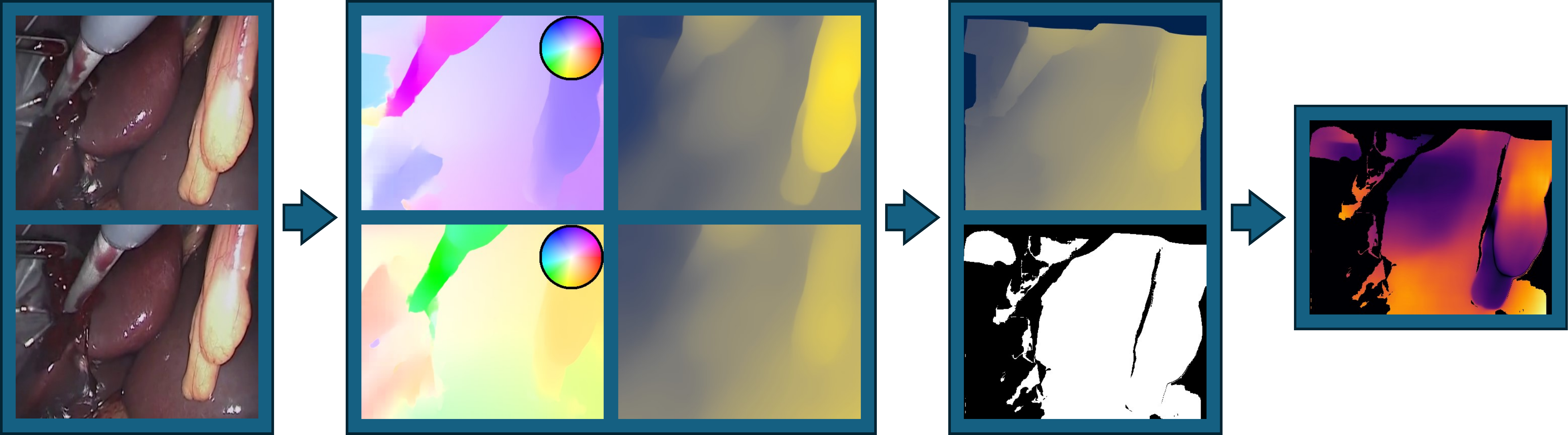}
\caption{Schematic of our temporal consistency loss. Starting from two temporally close input images at time $t$ and $t+\delta t$ (panel one), the optical flows and depth-maps are inferred (panel two). The flow is then used to calculate a correspondence mask, and to warp the inferred depth map at time $t+\delta t$ to align it with the image at time $t$ (panel three). The pixel-wise error between the warped and original depth map is masked (panel four) and averaged over to provide a final loss.}
\label{temploss}
\end{figure}

To address the temporal inconsistency we observed with pre-trained models,
we construct a loss which encourages the model to provide consistent results for
temporally close
frames.
We sample a pair of frames, $I_a$ and $I_b$, randomly from a clip ensuring the time between the frames does not exceed one tenth of a second.
The fast model and slow model are used to infer depth maps from $I_a$ and $I_b$ respectively.
We then compute bidirectional optical flow giving $F_{a\rightarrow b}$ and $F_{b\rightarrow a}$.
The depth map from $I_b$ is warped to align it with $I_a$ and masked with the correspondence mask calculated as in~\eqref{eq-mask} to provide a ground truth.
This process is illustrated in~\figref{temploss}.
The final temporal consistency loss may then be written as
\begin{equation}
    \mathcal{L}_{temp} = \SSIMAE\Big(M_{fast}(\vec{I_a}), F_{a\rightarrow b} \boldsymbol{\cdot} M_{slow}(\vec{I_b})\Big)
\end{equation}
We note that this construction ignores the effects of camera or subject motion along the optical axis. We hypothesis that enough positive supervision signal can be extracted during training before these errors become significant.

\subsubsection{Evaluation}
We evaluate 
performance using two criteria. 
We first re-purpose our $\SSIMAE$ metric to assess accuracy of the depth estimation on individual images.
The advantage of this metric is that it allows for direct comparison between datasets without worrying about changes in the shift or scale of the depth maps.
%
Then,
focusing on the errors which inspired our approach, we consider the temporal smoothness
throughout clips of endoscopic footage.

Using our correspondence masking, we identify sections of the SCARED Clip videos for which the majority of pixels in the start frame can be tracked to every frame.
We then use stereo disparity, and the models to be evaluated, to infer depth maps for all frames.
Depth maps from the monocular models are fitted to the start frame, as in~\eqref{eq-fit}, to account for shift and scale.
This allows us to build a “depth trajectory” of each tracked pixel throughout the clip.
We then subtract the stereo disparity trajectory from the inferred trajectories and calculate the standard deviation of the result.
This standard deviation is averaged over tracked pixels from all identified clips to provide a final metric.

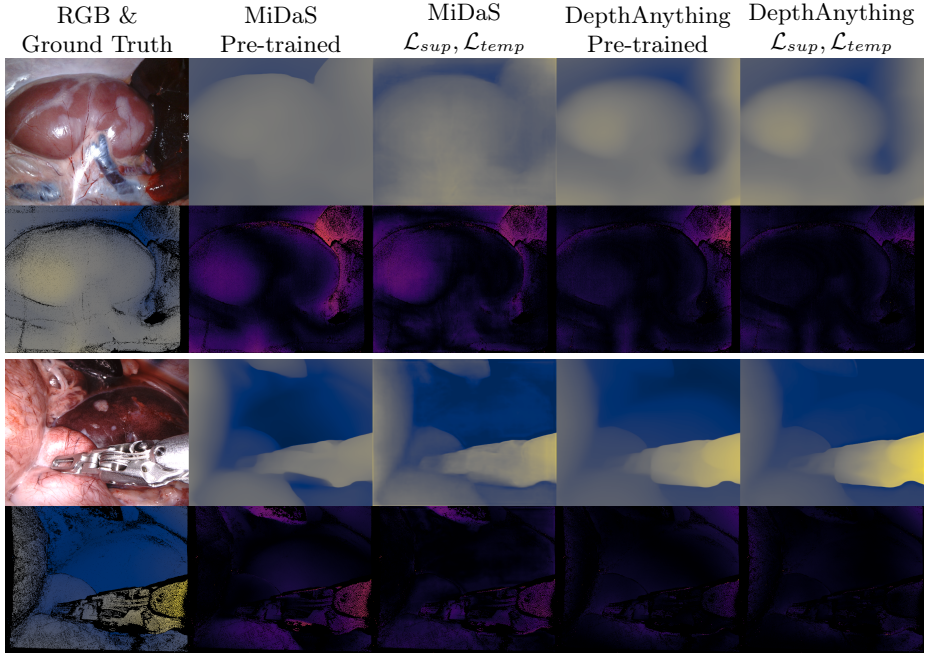
\begin{figure}
\centering
\begin{tikzpicture}
    \draw (-2*\textwidth/5, 2.35) node[align=center] {RGB \& \\ Ground Truth};
    \draw (-1*\textwidth/5, 2.35) node[align=center] {MiDaS \\ Pre-trained};
    \draw (0*\textwidth/5, 2.35) node[align=center] {MiDaS \\ $\mathcal{L}_{sup},\mathcal{L}_{temp}$};
    \draw (1*\textwidth/5, 2.35) node[align=center] {DepthAnything \\ Pre-trained};
    \draw (2*\textwidth/5, 2.35) node[align=center] {DepthAnything \\ $\mathcal{L}_{sup},\mathcal{L}_{temp}$};
    \draw (0, 0) node[inner sep=0] {\includegraphics[width=\textwidth]{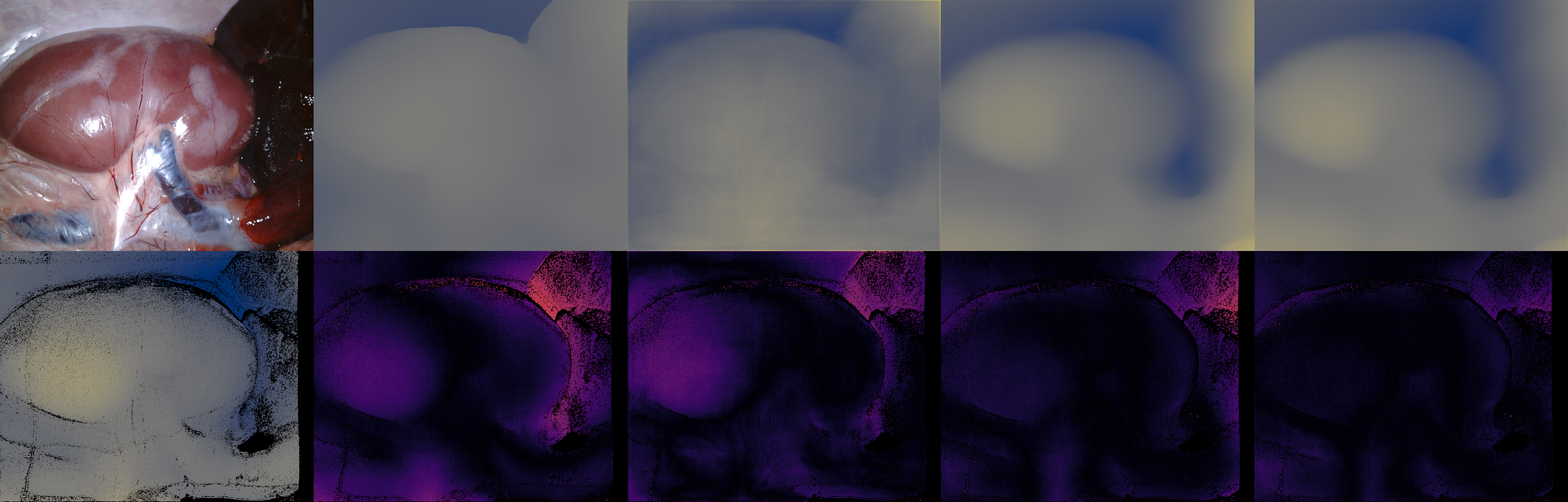}};
    \draw (0, -4) node[inner sep=0] {\includegraphics[width=\textwidth]{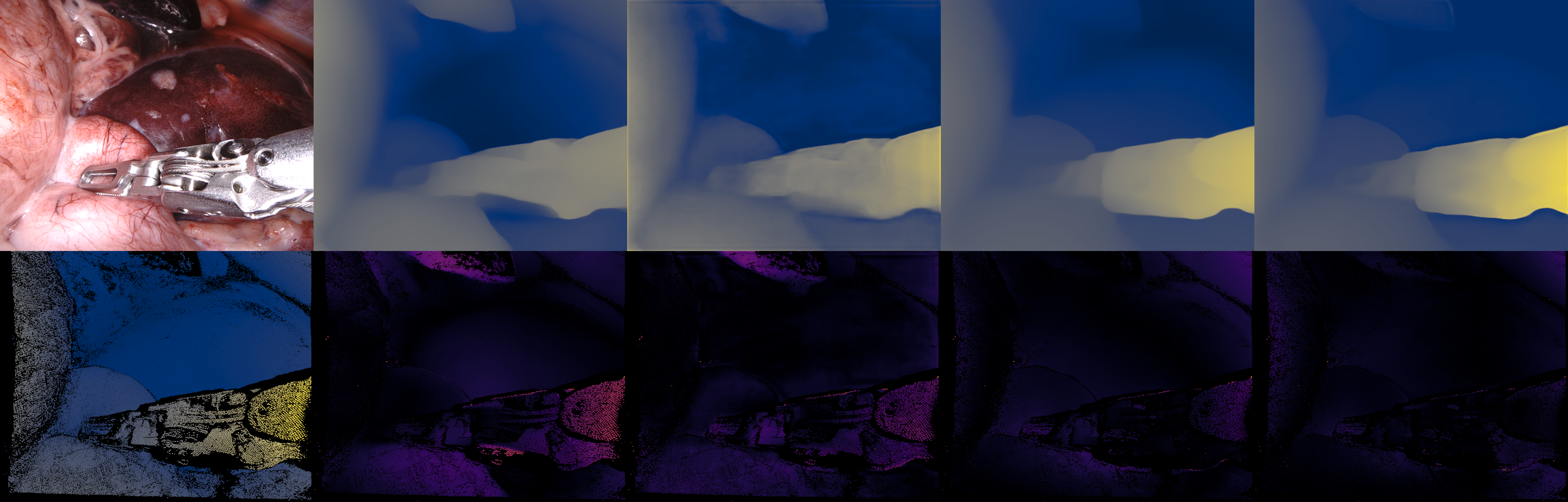}};
\end{tikzpicture}
\caption{Examples from our testing dataset. The left most column features the RGB (top) and the sparse ground truth depth (bottom), with the subsequent columns containing the predicted depth (top), which has been fitted to the ground truth as in~\eqref{eq-fit}, and the error (bottom) for a selection of models on a unified colour scale.}
\label{exampleoutputs}
\end{figure}

\begin{table}
\centering
\caption{Our scale and shift invariant MAE, and temporal inconsistency metrics for all our inference methods. The values shown for our trained models represent the mean and standard deviations, calculated across three training attempts. RAFT displays n/a for temporal inconsistency, as it is used in the construction of the metric.}
\label{errortable}
\begin{tabular}{|c|l|c|c|c|c|}
\hline

\multicolumn{2}{|c|}{\multirow{3}{*}{Method}} & \multicolumn{3}{|c|}{\multirow{2}{*}{Average SSIMAE}} & Temporal\\
\multicolumn{2}{|c|}{}&\multicolumn{3}{|c|}{}&Inconsistency\\

\cline{3-6}

\multicolumn{2}{|c|}{} & SCARED & SERV-CT & Both & SCARED Clips \\
\hline

\multicolumn{2}{|c|}{AF-SFMLearner~\cite{shao2022self}}
& 0.319* & 0.590 & 0.390* & 0.823* \\

\hline

\multicolumn{2}{|c|}{MiDaS}
& 0.370 & 0.418 & 0.382 & 1.266 \\

\hline

\multirow{4}{*}{MiDaS}
& $\mathcal{L}_{sup}$
& 0.355±.001 & \bftab{0.407±.001} & 0.369±.001 & 1.215±.003 \\

& $\mathcal{L}_{sup}, \mathcal{L}_{temp}$
& \bftab{0.316±.001} & 0.417±.001 & \bftab{0.343±.001} & \bftab{1.094±.003} \\

& $\mathcal{L}_{sup}, \mathcal{L}_{aug}$
& 0.317±.001 & 0.436±.002 & 0.349±.001 & 1.104±.002 \\

& $\mathcal{L}_{sup}, \mathcal{L}_{temp}, \mathcal{L}_{aug}$
& 0.333±.004 & 0.436±.009 & 0.360±.001 & 1.169±.012 \\

\hline

\multicolumn{2}{|c|}{DepthAnything}
& 0.309 & 0.342 & 0.318 & 1.102 \\

\hline

& $\mathcal{L}_{sup}$
& 0.280±.001 & 0.332±.002 & 0.294±.001 & 0.966±.005\\

Depth& $\mathcal{L}_{sup}, \mathcal{L}_{temp}$
& \bftab{0.265±.001} & 0.282±.004 & \bftab{0.269±.002} &\bftab{0.890±.014}\\

Anything& $\mathcal{L}_{sup}, \mathcal{L}_{aug}$
& 0.281±.004 & \bftab{0.280±.001} & 0.280±.003 & 0.924±.014\\

& $\mathcal{L}_{sup}, \mathcal{L}_{temp}, \mathcal{L}_{aug}$
& 0.268±.003 & 0.299±.015 & 0.276±.002 & 0.892±.009\\

\hline

\multicolumn{2}{|c|}{RAFT Stereo Disparity}
& 0.080  & 0.165 & 0.102 & n/a\\

\hline
\multicolumn{5}{l}{*Probable overfitting: Evaluation data observed during training.}
\end{tabular}
\end{table}

\section{Experiments}
For our fine-tuning experiments, we choose the best performing models from MiDaS and Depth Anything to act as our base models, specifically the MiDaS v3.1 dpt\_large\_512 and the Depth-Anything-Large models.
These transformer based models have very similar architectures and parameter counts ($345$ and $335$ million respectively), the main difference being the inclusion of more data, augmentation consistency self-supervision, and segmentation multitask learning in the training of the Depth Anything model. We fine-tune these base models using standard supervision, as well as semi-supervised learning using temporal consistency, augmentation consistency, and a combination of the two.
Images were prepared to match the preparations used in the original training of the base models, color jittered, and loaded in batches of $15$.
Models were optimised using SGD with a learning rate of $1$$\times$$10^{-6}$, and gradient norm clipping of $10$ to prevent collapse from unlucky batches during self-supervision.
Multiple objective functions are optimised for by using interleaved gradient updates for each loss~\cite{mayo2023multitask}.
We take $100$ batches to be an epoch and use our validation dataset for early stopping with a patience of $50$ epochs.

We present our error and smoothness metrics for all models in~\tabref{errortable}. Example outputs for selected models are shown in~\figref{exampleoutputs} and videos are available in the supplementary material.
The results show significant improvement over the base models from all fine-tuning methods.
Adding self-supervision offers a significant benefit to the training, more than doubling the reduction in error of standard supervision alone in almost all cases.
Temporal consistency proves to be the best self-supervision method for both base models, both in terms of SSIMAE and temporal consistency.
Finally we show clear superiority over  AF-SFMLearner, the domain-specific baseline.
Many of our models outperform this domain-specific baseline on SCARED despite the dataset being used to train the baseline.
When comparing performance on SERV-CT, data which is unseen for both methods, our best models show a reduction in error of over 50\%.

\section{Discussion and Conclusion}
In this work, we have demonstrated the ability of state-of-the-art transformer based relative monocular depth models, trained on huge natural image meta-datasets, to generalise reasonably to the surgical domain, and shown that careful fine-tuning can significantly improve the performance.
In line with the concurrent work of Depth Anything, we show that self-supervision can aid monocular depth models significantly.
Following the method used for Depth Anything, we also experimented with multitask learning for binary tool segmentation.
However, we found this to dramatically reduce performance, so full experiments were not run.
Furthermore, we show that our method of temporal consistency self-supervision significantly surpasses the predominantly used self-supervision method, augmentation consistency, when transferring to the surgical domain.
We have also demonstrate that our transferred models drastically outperform the state-of-the-art method developed specifically for endoscopy, demonstrating the
potential of transferring natural image models to endoscopy.
Finally, to aid in further research on this topic, we release our code, models, and dataset, \texttt{Meta-MED}, the meta monocular endoscopic depth dataset.

\anoncmd{
\subsubsection{Acknowledgements}
This study/project is funded by the NIHR [NIHR202114].
This work was supported by core funding from the Wellcome/EPSRC [WT203148/Z/16/Z; NS/A000049/1].
This project has received funding from the European Union's Horizon 2020 research and innovation program under grant agreement No 101016985 (FAROS project).
%
%
For the purpose of open access, the authors have applied a CC BY public copyright licence to any Author Accepted Manuscript version arising from this submission.
TV is a co-founder and shareholder of Hypervision Surgical.
}{}


\reffcmd
{
\printbibliography
}
{
\bibliography{library}{}
\bibliographystyle{splncs04}
}

\end{document}